\documentclass[sigconf]{acmart}

\usepackage{placeins}

\AtBeginDocument{%
  }

\copyrightyear{2026}
\acmYear{2026}
\setcopyright{cc}
\setcctype{by}
\acmConference[KDD '26]{Proceedings of the 32nd ACM SIGKDD Conference on Knowledge Discovery and Data Mining V.2}{August 09--13, 2026}{Jeju Island, Republic of Korea}
\acmBooktitle{Proceedings of the 32nd ACM SIGKDD Conference on Knowledge Discovery and Data Mining V.2 (KDD '26), August 09--13, 2026, Jeju Island, Republic of Korea}
\acmDOI{10.1145/3770855.3818463}
\acmISBN{979-8-4007-2259-2/2026/08}

\begin{document}

\title{Pinterest Canvas: Large-Scale Image Generation at Pinterest}

\author{Yu Wang}
\email{yuw@pinterest.com}
\affiliation{
  \institution{Pinterest, Inc.}
  \country{San Francisco, CA, USA}
}

\author{Eric Tzeng}
\email{etzeng@pinterest.com}
\affiliation{
  \institution{Pinterest, Inc.}
  \country{San Francisco, CA, USA}
}

\author{Raymond Shiau}
\email{rshiau@pinterest.com}
\affiliation{
  \institution{Pinterest, Inc.}
  \country{San Francisco, CA, USA}
}

\author{Jie Yang}
\email{jieyang@pinterest.com}
\affiliation{
  \institution{Pinterest, Inc.}
  \country{San Francisco, CA, USA}
}

\author{Dmitry Kislyuk}
\email{dkislyuk@pinterest.com}
\affiliation{
  \institution{Pinterest, Inc.}
  \country{San Francisco, CA, USA}
}

\author{Charles Rosenberg}
\email{crosenberg@pinterest.com}
\affiliation{
  \institution{Pinterest, Inc.}
  \country{San Francisco, CA, USA}
}

\begin{abstract}
While recent image generation models demonstrate a remarkable ability to handle a wide variety of image generation tasks, this flexibility makes them hard to control via prompting or simple inference adaptation alone, rendering them unsuitable for use cases with strict product requirements.
In this paper, we introduce \textit{Pinterest~Canvas}, our large-scale image generation system built to support image editing and enhancement use cases at Pinterest.
Canvas is first trained on a diverse, multimodal dataset to produce a foundational diffusion model with broad image-editing capabilities.
However, rather than relying on one generic model to handle every downstream task, we instead rapidly fine-tune variants of this base model on task-specific datasets, producing specialized models for individual use cases.
We describe key components of Canvas and summarize our best practices for dataset curation, training, and inference.
We also showcase task-specific variants through case studies on background enhancement and aspect-ratio outpainting, highlighting how we tackle their specific product requirements.
Online A/B experiments demonstrate that our enhanced images receive a significant 18.0\% and 12.5\% engagement lift, respectively, and comparisons with human raters further validate that our models outperform third-party models on these tasks.
Finally, we showcase other Canvas variants, including multi-image scene synthesis and image-to-video generation, demonstrating that our approach can generalize to a wide variety of potential downstream tasks.
\end{abstract}



\begin{teaserfigure}
  \centering
  \includegraphics[width=1.0\textwidth]{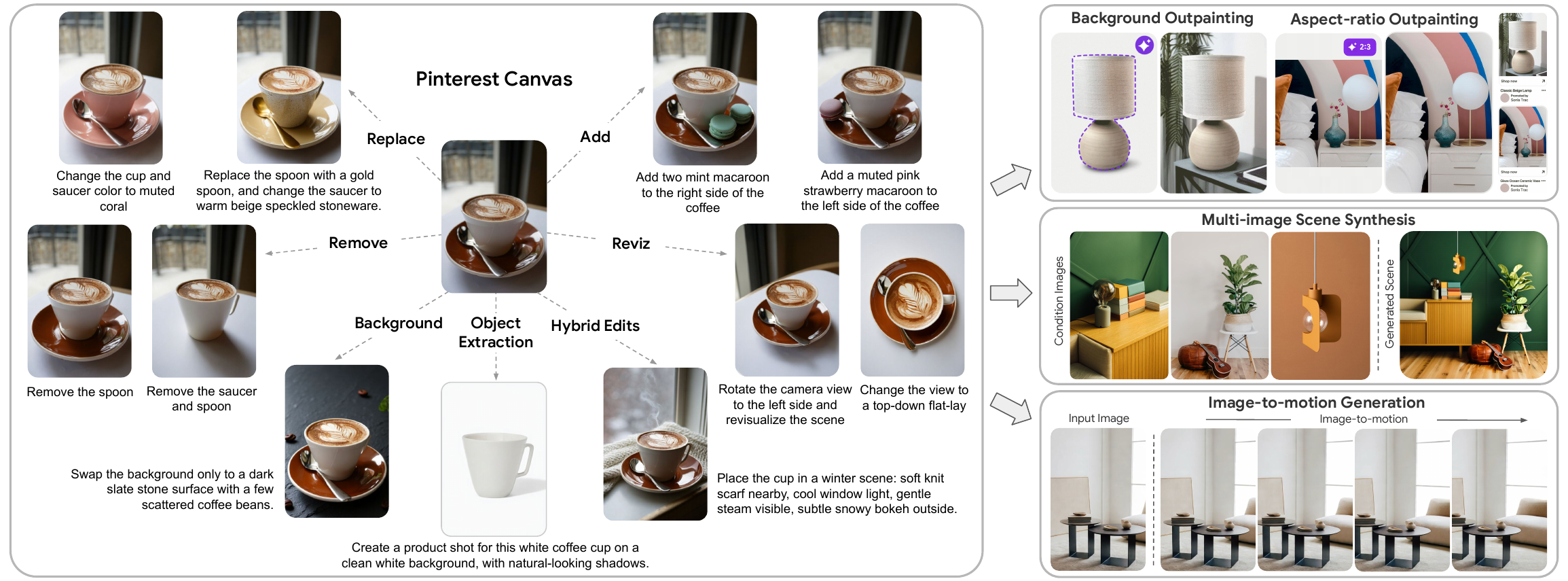}
  \caption{
  Rather than relying solely on a single generic image generation model to handle every potential task, in our design of \textit{Pinterest Canvas} we opt to rapidly fine-tune dedicated variants that are tailored to each use case's unique product requirements.
  }
  \Description{
  A diagram showcasing our base model's ability to handle flexible yet generic image editing tasks, as well as three example task-specific variants: image enhancement for advertising, multi-image scene synthesis, and image-to-motion generation.
  }
  \label{fig:teaser}
\end{teaserfigure}

\maketitle

\section{Introduction}

Ever since the introduction of diffusion models~\cite{ho2020denoisingdiffusionprobabilisticmodels, song2021scorebasedgenerativemodelingstochastic, rombach2022highresolutionimagesynthesislatent}, the quality and capabilities of image generation models have rapidly and dramatically improved over the last few years.
Pinterest has a rich ecosystem of high quality visual data, making us well-suited for being able to leverage this technology to further enhance the content we currently serve to users.

However, users of Pinterest often browse with concrete intent, searching for ideas and visuals that they may want to introduce into their lives, such as fashion looks or home decor inspiration.
This means that na\"ive applications of these models, such as simply generating results for search queries, risk detracting from the product---synthesized results may not correspond to anything that exists or can be acted upon, and therefore may fail to satisfy user intent.
Thus, we have found that it often makes sense to limit applications of these models to revisualization and enhancement of existing content, rather than wholesale generation of new imagery.

Our emphasis on seamless editing and enhancement imposes additional requirements upon our models.
We often need to balance broad, sweeping edits like background generation against careful preservation of real objects and products.
Commonly available models~\cite{esser2024scalingrectifiedflowtransformers, labs2025flux1kontextflowmatching, wu2025qwenimagetechnicalreport, gptimage, nanobanana} are often positioned as general image generation or enhancement models controlled primarily via prompting, making them incredibly versatile, but also leaving them poorly positioned to support these requirements.

Additionally, the requirements imposed upon each individual revisualization task can directly contradict each other---as a concrete example, generating new backgrounds for products requires strict preservation of the original product, whereas insertion of products into scenes often necessitates altering the pose of the product in order to match the perspective of the target scene.
This makes it difficult to rely on a single generic model to handle every potential downstream task.

As a result, we devise a framework that trains one foundational model, then finetunes dedicated variants of that model for each revisualization task we need to satisfy.
This family of models, which we call Pinterest Canvas, incurs the cost of training a foundational diffusion model once upfront, but then allows us to rapidly produce descendant models that are targeted for specific use cases and requirements, as shown in Figure~\ref{fig:teaser}.
Additionally, controlling the training of this model at the foundational stage allows us to pretrain the model on a variety of tasks related to the final intended uses, thereby speeding up convergence when producing task-specific variants. Furthermore, this end-to-end control also applies to the training dataset, allowing us to curate inputs and ensure safety, compliance, quality, and relevance, which in turns enables us to ship models with confidence.

In this paper, we provide a comprehensive technical overview of Pinterest Canvas.
We describe the design, dataset, and training strategies used to produce our foundational base model, which can flexibly handle a wide variety of image generation and editing tasks.
As a demonstration of our strategy for quickly training product-focused variants, we provide a case study of our Canvas models for ads enhancement, outlining the fine-tuning procedure, describing the various extensions we implemented to improve performance in production settings, and providing offline and online evaluations to validate performance. Finally, to further validate our overall framework, we showcase two other Canvas variants: multi-image scene synthesis and image-to-motion video generation.

\section{Related Work}
Diffusion models~\cite{ho2020denoisingdiffusionprobabilisticmodels, song2021scorebasedgenerativemodelingstochastic, rombach2022highresolutionimagesynthesislatent} have revolutionized image generation, with substantially improved image quality.
Diffusion Transformers (DiTs)~\cite{peebles2023scalablediffusionmodelstransformers} replace the U-Net backbone with transformer blocks and have demonstrated strong image generation quality and scaling capability. 
Stable Diffusion 3~\cite{esser2024scalingrectifiedflowtransformers} combines a multimodal DiT (MM-DiT) architecture with rectified flow~\cite{lipman2023flowmatchinggenerativemodeling} to achieve strong text-to-image generation performance. 
FLUX.1~\cite{blackforestlabs_flux1_2024} further adopts a double-stream MM-DiT architecture and rotary positional embeddings (RoPE), improving both image quality and text coherence.

Text-to-image generation is a crucial foundation for image synthesis, but most real-world scenarios require images as conditions. Prior work typically adds this capability through auxiliary modules~\cite{hu2021loralowrankadaptationlarge, zhang2023addingconditionalcontroltexttoimage, ye2023ipadaptertextcompatibleimage} or task-specific fine-tuning~\cite{gal2022imageworthwordpersonalizing, ruiz2023dreamboothfinetuningtexttoimage}. FLUX.1 Kontext~\cite{labs2025flux1kontextflowmatching} encodes conditioning images via VAE and concatenates them with image tokens, enabling unified in-context image conditioning without modifying the base architecture. Qwen-Image~\cite{wu2025qwenimagetechnicalreport} adopts a dual-path conditioning scheme to capture both high-level intent and to provide fine-grained visual semantics.

In parallel, autoregressive models have demonstrated strong results in image generation~\cite{yu2022scalingautoregressivemodelscontentrich, tian2024visualautoregressivemodelingscalable}. Several proprietary commercial systems also show strong performance and likely benefit from their broader LLM ecosystems, such as OpenAI’s GPT-Image~\cite{gptimage} and Google’s Nano Banana~\cite{nanobanana}. 

\paragraph{Outpainting}
Image outpainting extends an image beyond its original boundaries with new content. 
Early diffusion-based outpainting methods~\cite{saharia2022paletteimagetoimagediffusionmodels, yang2024vipversatileimageoutpainting} operate within a rectangular generation window, restricting their scaling to arbitrarily large expansions.
Recent works improve controllability via explicit conditioning and prevent foreground outpainting by conditioning on instance masks, but are often trained on web data~\cite{alimama2024ecomxl_controlnet_inpaint, alimama2024flux_controlnet_inpainting} or synthetic data~\cite{zhao2025dreampainterimagebackgroundinpainting}, which may generalize poorly to real-world use cases.

\paragraph{Scene Synthesis}
Interest has recently grown in flexible scene synthesis with one or more specified subjects for richer visualization.
Prior work~\cite{ye2023ipadaptertextcompatibleimage,zeng2024jedijointimagediffusionmodels} typically injects a single reference via a pretrained image encoder or VAE with attention mechanisms, which can improve identity preservation and novel-view generation but is often limited to one subject and becomes costly when jointly denoising multiple images. 
Other lines of work~\cite{li2025homediffusion,fu20213dfront3dfurnishedrooms,wang2025roomeditorparametersharingdiffusionarchitecture} explore multi-view revisualization and object/background editing using synthetic 3D data or parameter-sharing diffusion backbones, but they can suffer from limited real-world realism and are frequently restricted to basic mask conditioning and single-reference insertion into a fixed background.

\paragraph{Image-to-motion Generation}
Turning a static image into a short motion clip is another extension of product image enhancement. The rapid evolution of video generation models provides one path to this application. 
Recent work~\cite{wan2025, opensora2} proves the feasibility of conditioning diffusion backbones on reference images for image-to-video generation, but these models are typically optimized for longer narrative-style videos and rely on large-scale text-to–video pretraining. 
Another line of work~\cite{guo2023animatediff} reuses text-to-image diffusion models for motion by injecting lightweight motion modules, enabling efficient adaptation of short motion synthesis, but often with a narrower motion range and occasional temporal artifacts.

\section{System Overview}
As previously motivated, in order to ensure our models can consistently satisfy the unique requirements of each individual downstream task, instead of relying on a single generic model, we want to train dedicated models that target each use case and directly incorporate the aforementioned requirements.
Of course, training a dedicated model from scratch for each task is inefficient and wasteful.
Our design for Pinterest Canvas strikes a compromise between the two approaches.

First, we train a large-scale, generic model to perform a variety of image generation and editing tasks.
However, unlike the monolithic approach that directly applies this generic model to every downstream task, we instead use it as a base model for multiple task-specific variants that are further fine-tuned on dedicated datasets.
By training this base model across a broad spectrum of generation tasks, we ensure that it serves as a strong set of base weights, allowing any descendant models to rapidly converge when trained on focused datasets.
Conversely, because these descendant models are able to focus on a single targeted editing operation, their behavior is much easier to control, since we are able to limit training samples to only those which exhibit the desired properties.

We begin with a discussion of our dataset, describing how it is collected and what tasks we incorporate in order to ensure that our foundational model is well-suited to support a variety of use cases.
Next, we describe the training protocol for our base model, providing details on architecture and optimization.
Finally, we provide a high-level overview of how this base model can be fine-tuned on focused datasets to produce variant models that are specialized for their intended purposes.

\subsection{Dataset}
Pinterest hosts a large corpus of high-quality visual content. To train a large-scale image generation model, we systematically collect and annotate billions of paired text--image and (text+image)--image data. We apply aggressive filtering to ensure compliance with user opt-outs, strong visual quality, and high relevance between the target content and its conditioning signals.

\begin{figure*}[t]
  \centering
  \includegraphics[width=0.98\linewidth]{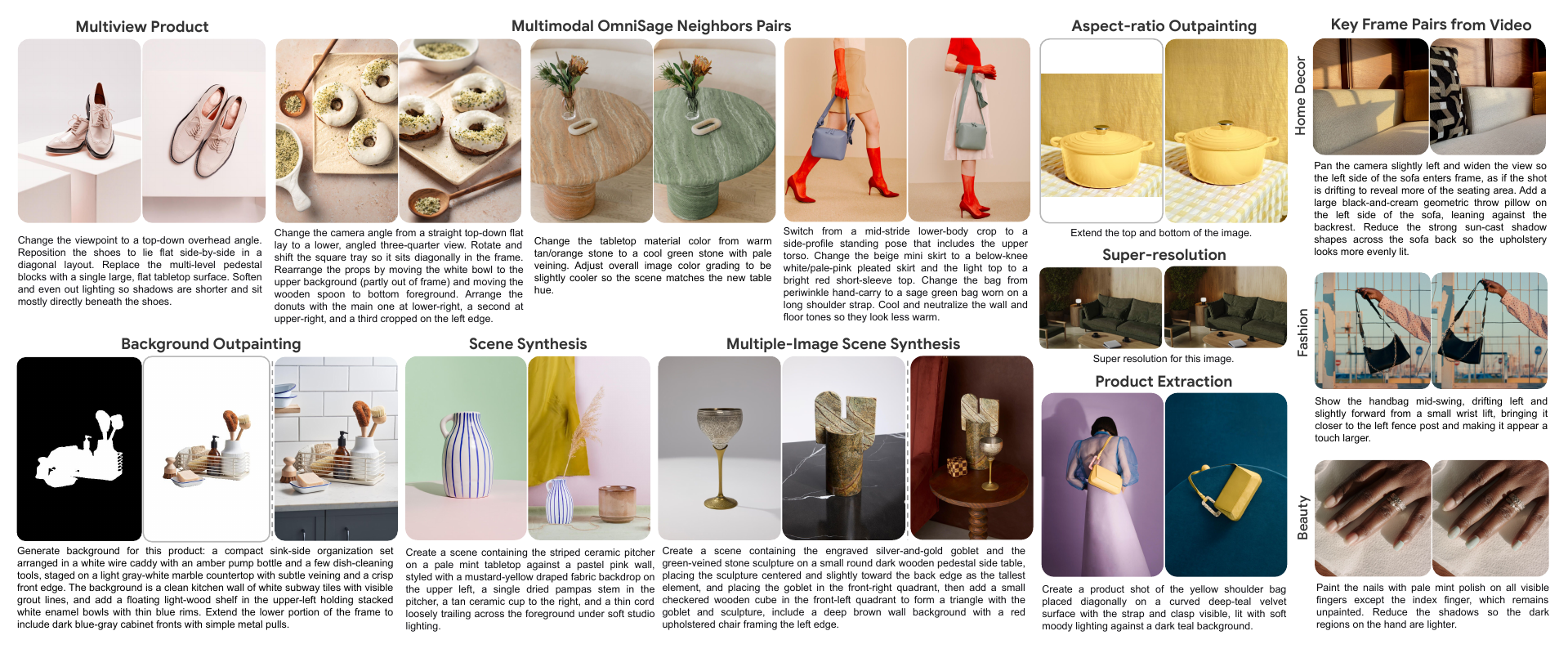}
  \caption{Multimodal dataset visualization across tasks. In each example, the left image(s) is the input condition(s), and the right image is the target output; the text below each pair describes the transformation required to produce the target from the input.}
  \label{fig:dataset}
\end{figure*}

\subsubsection{Text--Image Data}
We begin with more than 17 billion source-compliant text--image pairs and apply additional filtering for text--image relevance, text quality, and image quality. We use VLMs to generate high-quality captions at varying lengths. Even after this stringent filtering, we retain more than 2.6 billion high-quality text–image pairs, providing sufficient scale for a long training schedule and enabling Canvas to converge to high-quality image generation with an appealing aesthetic.

\subsubsection{(Text+Image)--Image Data}
Beyond generic text--image pairs, we also curate a suite of multimodal datasets for image editing. These datasets are built through multiple pipelines and are designed to cover both general editing capabilities and product-specific use-cases. Visualization examples of each task are shown in Figure~\ref{fig:dataset}.

\paragraph{Multi-view product.}
This data consists of sets of images from the same item to support multi-view revisualization with strong identity preservation across viewpoint changes.

\paragraph{OmniSage neighbors data.}
This data is structured as (anchor Pin image + instruction) $\rightarrow $ neighbor Pin image, where the neighbor is retrieved from OmniSage~\cite{Badrinath_2025} and the instruction describes the semantic edit needed to transform the anchor into the neighbor.

\paragraph{Background outpainting.}
Each example includes a lifestyle product image paired with a salient-object mask segmented from InSPyReNet~\cite{kim2022revisitingimagepyramidstructure} and a caption describing the product and surrounding background.

\paragraph{Aspect-ratio outpainting.}
We synthesize this dataset from text--image data by applying random height masking to the original image and masking out the top and bottom regions to be generated.

\paragraph{Super-resolution.}
Each example is a synthetic low-/high-resolution pair for $\times3$ restoration, generated by applying ESRGAN degradations~\cite{wang2018esrganenhancedsuperresolutiongenerative} to images from our text--image dataset.

\paragraph{Product extraction and scene synthesis.}
Each example is a matched pair of product shot and a scene containing that product. Swapping input/target yields two training formats: scene $\rightarrow$ product (product extraction) and product $\rightarrow$ scene (scene synthesis).

\paragraph{Multi-image scene synthesis.}
This data consists of a scene paired with a set of one or more product images (up to eight images) in that scene. We caption the scenes, describing both the overall setting and the placement of prominent objects.

\paragraph{Video-augmented key frame pairs.}
We augment training with key-frame pairs from filtered high-quality video clips, formatted as $\text{frame}_t\rightarrow\text{frame}_{t+\Delta}$ with $\Delta\ge2$s, to learn temporally consistent transformations.

\vspace{0.5em}
All captions are generated with VLMs from the input image(s), and most datasets additionally use VLM-based quality control. Full dataset construction details are provided in Appendix~\ref{app:details_training_dataset}.

\subsubsection{Multiple-task Joint Learning}
We mix datasets and tune the dataset weighting to balance tasks during training, and we observe clear benefits from joint learning for the base model. Many tasks exhibit positive transfer and reinforce one another, leading to faster learning than training on a single task, while enabling a single unified model architecture to support a wide range of image editing use cases.
The multimodal dataset is composed of the following task types with ratios: multi-view product data 27.0\%, OmniSage neighbors 22.2\%, video-augmented keyframe pairs 14.2\%, background outpainting 11.7\%, aspect-ratio outpainting 10.7\%, super-resolution 7.1\%, multi-image scene synthesis 3.5\%, product extraction 1.8\%, and scene synthesis 1.8\%. 
The majority of the data consists of samples that are readily available from the corpus or from prior work; the remaining smaller-scale datasets were collected specifically to support expected downstream applications.
To help the model distinguish between tasks and better generate outputs that meet each specific set of requirements, we prepend task-specific prefixes to the text captions, such as ``\textit{Generate background for this product:}'' for background outpainting, ``\textit{Extend the top and bottom of the image}'' for aspect-ratio outpainting, and ``\textit{Super resolution for this image}'' for super-resolution.

\subsubsection{User Opt-out Compliance}
As part of our efforts to respect user privacy settings, we regularly update our training datasets to account for changes in user privacy settings or training opt-out requests.
The only way to guarantee that the model is not influenced by these data points is to ensure that it is never trained on this data---thus, 
we regularly retrain our models from scratch, and the training data used in each run reflects the most recent state of these settings.
This is yet another advantage of training the Canvas family of models entirely ourselves: it allows us to provide strong guarantees to our users about our responsible use of data, and can further mitigate the risk of potentially producing unsafe generations.

\subsection{Model Training}
Our goal is to train a model that is able to generate images conditioned on both text prompts and reference images. Condition images are encoded by the VAE encoder and are appended to the latent tokens via sequence concatenation. We adopt the FLUX.1 Kontext~\cite{labs2025flux1kontextflowmatching} backbone, combining double-stream DiT blocks with single-stream DiT blocks, and the model is trained via flow matching.                                                                                                           
We optimize with AdamW using $\beta_2 = 0.95$ and gradient clipping with a threshold of 4. We also apply 10\% text dropout and 10\% image dropout during training to enable classifier-free guidance at inference.
To improve both training efficiency and model performance, we use a multi-stage training strategy.
In the first stage, we train the model from scratch on text-to-image generation at an average resolution of $256^2$, with bucket sampling for varying aspect ratios. In the second stage, we resume training with multimodal image editing tasks at the $256^2$ scale. In subsequent stages, we further scale multimodal training to higher resolutions of $512^2$ and $1024^2$.

We find that training multimodal image editing tasks at lower resolutions not only speeds up model convergence at higher resolutions, but also improves the image fidelity and identity preservation of the final model, as compared to training on the multimodal image editing dataset directly at $512^2$ in the second stage.

\subsubsection{Timestep Shifting}
\begin{figure}[t]
  \centering
  \includegraphics[width=0.95\linewidth]{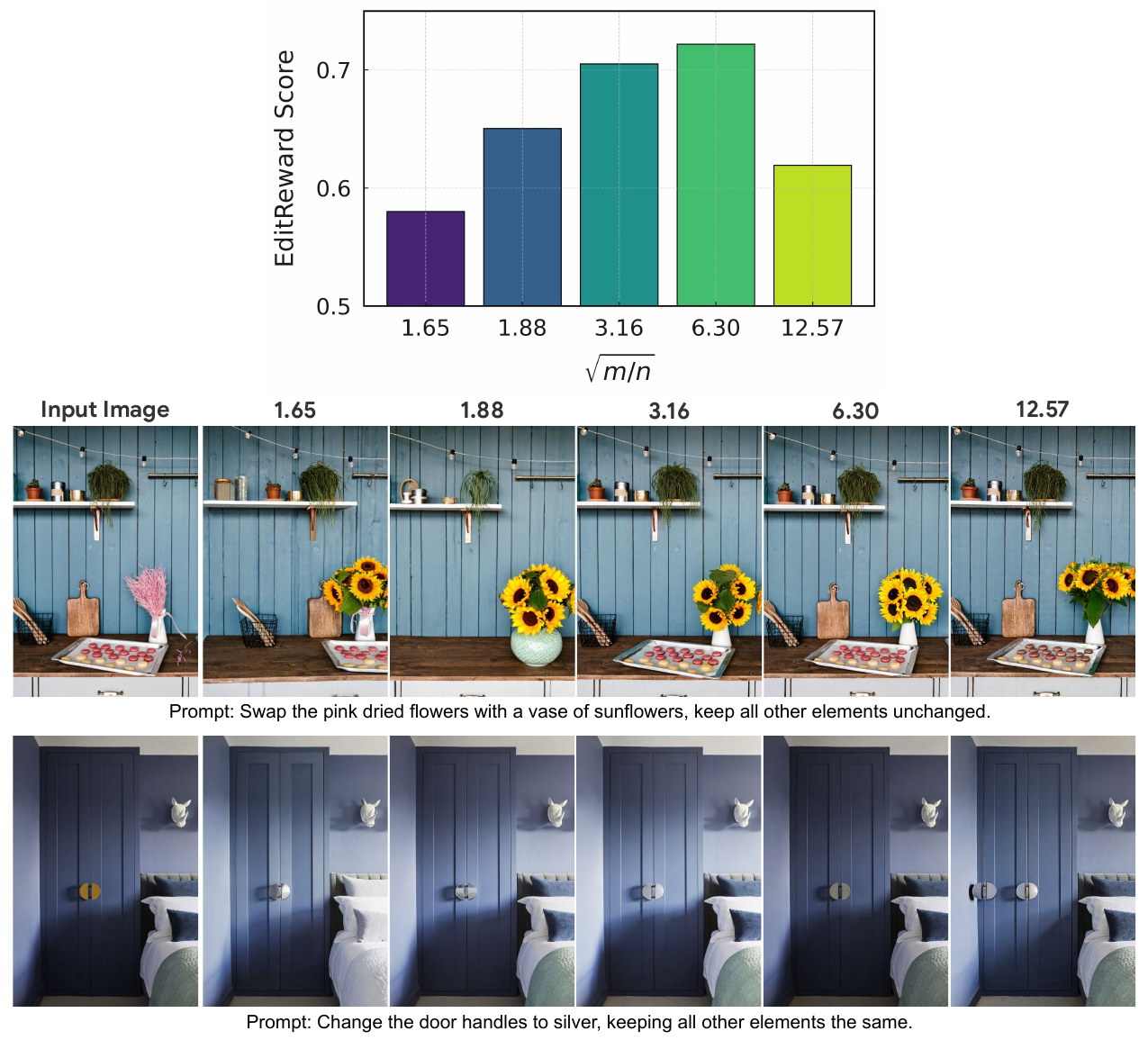}
  \caption{Timestep shifting experiments. Averaged EditReward scores and qualitative examples at each $\sqrt{m/n}$ setting.}
  \Description{Timestep shifting}
  \label{fig:timestep_shifting}                                                       
\end{figure}

Since higher resolution images need more noise to destroy the signal, we apply the timeshift transform
\begin{equation}
t_m \;=\;
\frac{\sqrt{\frac{m}{n}}\, t_n}{1 + \left(\sqrt{\frac{m}{n}} - 1\right)t_n},
\label{eq:timestep_shift}
\end{equation}
where $\sqrt{m/n}$ controls the sampling timeshift. This formula maps a timestep $t_n$ at resolution $n$ to a corresponding timestep $t_m$ at resolution $m$ such that the two settings have the same degree of uncertainty. Prior work~\cite{esser2024scalingrectifiedflowtransformers, BFL2025RepresentationComparison} has shown this timeshift to be effective for text-to-image training. 
Here, we further evaluate its applicability and empirically identify a good setting for multimodal training. At the $512^2$ scale, we sweep $\sqrt{m/n} \in \{1.65,\;1.88,\;3.16,\;6.30,\;12.57\}$ and evaluate on an internal image editing benchmark of 272 manually curated text+image input pairs, covering both general image editing tasks and product focused use-cases. For each sample, we generate four outputs per input with different seeds to reduce the impact of sampling randomness (1,088 evaluated images per setting), and score them with the EditReward model~\cite{wu2025editrewardhumanalignedrewardmodel}, a human-aligned reward model for instruction-guided image editing. In this experiment, $\sqrt{m/n} = 6.30$ yields the best overall performance both quantitatively by average EditReward scores and qualitatively, producing higher-fidelity edits with better identity preservation and fewer hallucinations, as shown in Figure~\ref{fig:timestep_shifting}. This aligns with the optimal shift previously reported~\cite{esser2024scalingrectifiedflowtransformers} for text-to-image training at the same resolution.  

\subsubsection{Training Stability}
Training stability is critical and challenging for large-scale and high-resolution diffusion models.
In our runs, we observed occasional loss spikes early in training or mid-run.
When these spikes occurred, model quality often suffered catastrophically and permanently---even if the loss later returned to its previous levels, we found that output quality was noticeably worse.
We also observed slow convergence and occasional divergence in longer runs.
To improve stability, we experimented with several strategies, including lowering the learning rate, applying gradient clipping, tuning optimizer hyperparameters, incorporating magnitude-preserving operations~\cite{karras2024analyzingimprovingtrainingdynamics}, and maintaining an exponential moving average (EMA) of the model weights.
Despite our many attempts and experiments, ultimately we found that the most impactful interventions were simply decreasing the exponential decay rate for the second-moment estimates $\beta_2$ in AdamW to 0.95~\cite{wortsman2023stablelowprecisiontraininglargescale} and enabling EMA.

\subsection{Multimodal Classifier-Free Guidance}
Classifier-free guidance (CFG) is defined as~\cite{ho2022classifierfreediffusionguidance}
\begin{equation}
x_{\mathrm{cfg}}
=
\epsilon_{\theta}(z_t,\varnothing)
+ s \cdot \left(\epsilon_{\theta}(z_t,c_T)-\epsilon_{\theta}(z_t,\varnothing)\right),
\label{eq:cfg}
\end{equation}
where $s$ is a hyperparameter controlling guidance scale, $z_t$ denotes the noisy latent at diffusion timestep $t$, and $\epsilon_\theta(\cdot)$ is the model prediction when the conditioning text is provided or dropped out.
However, the original formulation only handles the case where we have a single text caption.
When we generalize from one condition to multiple (e.g. a text condition $c_T$ and an image condition $c_I$), the number of ways we might apply CFG grows combinatorially, since each condition can be individually kept or dropped.
Prior work on multi-condition guidance imposes an ordering on the conditions, then applies CFG on each successive condition as it is added back in~\cite{brooks2023instructpix2pixlearningfollowimage}:
\begin{equation}
\begin{aligned}
x_{\mathrm{cfg}}^{(I,T)}
&=\epsilon_{\theta}(z_t,\varnothing,\varnothing) \\
&\phantom{=} \  + s_I \cdot \left(\epsilon_{\theta}(z_t,\varnothing,c_I)-\epsilon_{\theta}(z_t,\varnothing,\varnothing)\right) \\
&\phantom{=} \  + s_T \cdot \left(\epsilon_{\theta}(z_t,c_T,c_I)-\epsilon_{\theta}(z_t,\varnothing,c_I)\right).
\end{aligned}
\label{eq:cfg_multi}
\end{equation}
While this type of CFG allows us to easily trade off the contribution of each input condition, it still incurs an additional forward pass in each iteration, with the number of additional passes growing linearly as the number of input conditions increases.

In order to maintain inference efficiency, we simplify multi-condition CFG to one of two variants, both of which only require two forward passes per iteration:
\begin{align}
&x_{\mathrm{cfg}}^{(T)}
&=
\epsilon_{\theta}(z_t,\varnothing,c_I)
+ s \cdot \left(\epsilon_{\theta}(z_t,c_T,c_I)-\epsilon_{\theta}(z_t,\varnothing,c_I)\right),
\label{eq:cfg_uncond_prompt}
\\
&x_{\mathrm{cfg}}^{(T+I)}
&=
\epsilon_{\theta}(z_t,\varnothing,\varnothing)
+ s \cdot \left(\epsilon_{\theta}(z_t,c_T,c_I)-\epsilon_{\theta}(z_t,\varnothing,\varnothing)\right),
\label{eq:cfg_uncond_prompt_image}
\end{align}
Qualitatively, we found that the variant in Eq.~\eqref{eq:cfg_uncond_prompt} adheres more closely to the prompt and yields higher overall image fidelity.
In contrast, the variant in Eq.~\eqref{eq:cfg_uncond_prompt_image} better preserves the reference image, though it can occasionally produce oversaturated outputs when using large guidance scales.
In our experiments, we found that these simplified variants produced outputs comparable to the complete formulation in Eq.~\ref{eq:cfg_multi} while running in two-thirds the time.

We observed that the best-performing CFG variant and guidance scale varied depending on the task.
For example, Eq.~\eqref{eq:cfg_uncond_prompt} works best for general image editing and background outpainting with an optimal scale of $s=7$, while Eq.~\eqref{eq:cfg_uncond_prompt_image} performs better for aspect-ratio outpainting at $s=3$.
For super-resolution we use $s=1$, which corresponds to disabling CFG outright.
Also, we found that judicious use of negative prompts can help suppress hallucinations and artifacts for many of our image-editing tasks, but for tasks where the input conditions strongly constrain plausible outputs, such as aspect-ratio outpainting, we observed that an empty negative prompt produced better results.

Finally, prior work~\cite{lin2024commondiffusionnoiseschedules} suggests rescaling outputs produced via CFG based on the original conditional generation, e.g. for Eq.~\ref{eq:cfg_uncond_prompt}:
\begin{equation}
x_{\mathrm{cfg}\text{-}\mathrm{norm}}^{(T)}
=
x_{\mathrm{cfg}}^{(T)}
\cdot
\frac{\operatorname{std}\!\left(\epsilon_\theta(z_t, c_T, c_I) \right)}{\operatorname{std}\!\left(x_{\mathrm{cfg}}^{(T)}\right)}.
\label{eq:noise_normalization}
\end{equation}
This can help mitigate overexposure issues that occur when using large guidance scales.
We adopt this strategy during inference, combining it with our choice of CFG variant, and observe that in multimodal conditioning settings, this rescaling has the additional benefit of improving adherence to the reference image.

\subsection{Fine-tuning for Downstream Tasks}
The base Canvas model is trained generically to provide a strong pretrained foundation, but task-specific fine-tuning on dedicated datasets allows variants to focus on a single target task and better meet product needs without optimizing for unrelated objectives. 
The flexible conditioning of the base model also makes it relatively straightforward to train task-specific variants by only switching the training data while keeping the same model architecture.

Since base training already covers the targeted tasks or relevant pretraining tasks, it provides a strong starting point for specialization and enables rapid fine-tuning. For example, outpainting tasks are included in base training with the same input format, single-image scene synthesis supports multi-image scene synthesis, and paired images with temporally consistent transformations support image-to-motion generation. 
Overall, this strategy makes it easy to incorporate future use cases into base Canvas training and to fine-tune efficiently for downstream applications.

\section{Canvas for Ads Enhancement}
\begin{figure*}[t]
  \centering
  \includegraphics[width=0.93\linewidth]{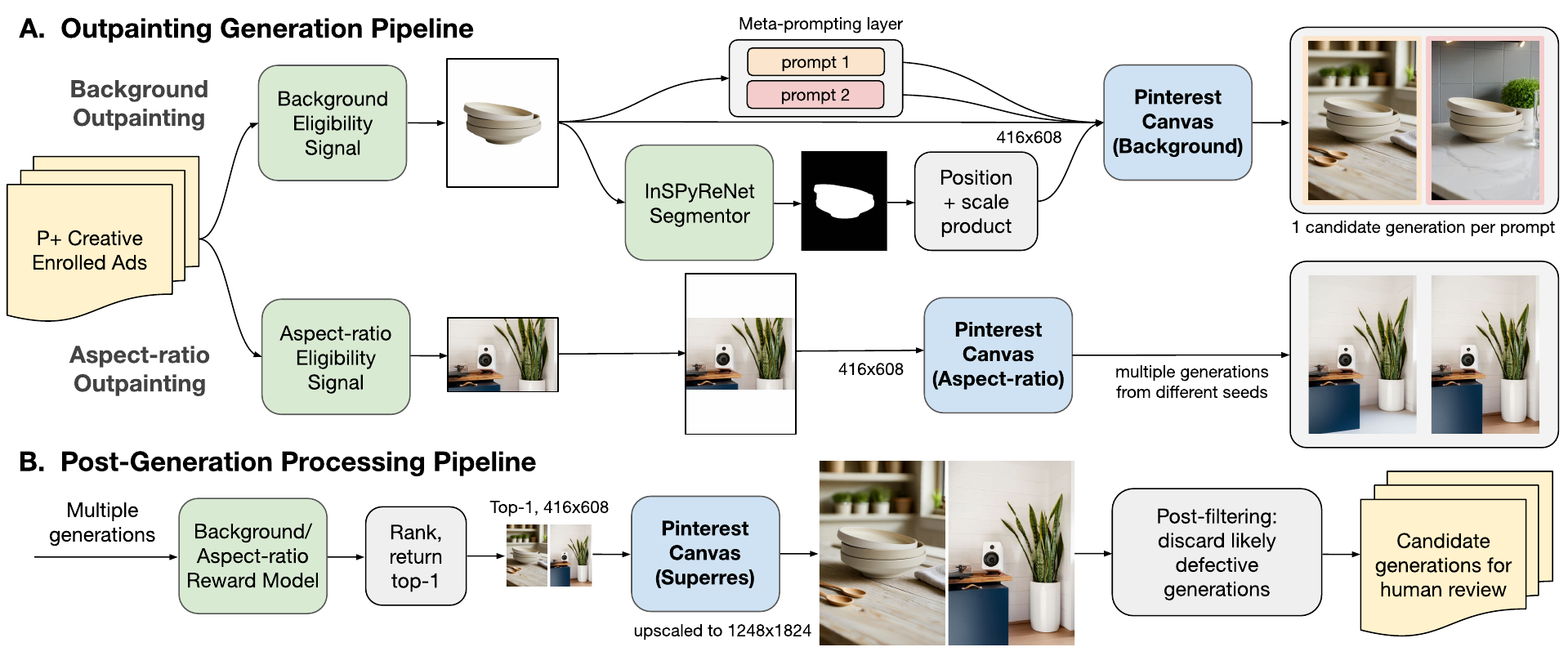}
  \caption{Canvas outpainting inference pipeline.}
  \Description{Canvas outpainting inference pipeline diagram}
  \label{fig:outpainting_inference_diagram}
\end{figure*}

We now discuss in detail one such example of a task-specific Canvas variant that we have deployed in production.
This model is offered as part of the Pinterest Performance+ suite of tools provided to advertising partners~\cite{pinterest_performance_plus}. 

We focus on two main categories of image enhancements.
The first is automatic background generation, which can quickly help advertisers generate more engaging backgrounds for any plain, white background photos in their catalog.
The second is aspect ratio expansion, which takes square or landscape photos and extends them vertically---due to Pinterest's columnar layout, taller images often perform better, so this tool is intended to help advertisers quickly modify their existing imagery into an aspect ratio that is better suited to the site.

While general-purpose models are often capable of performing these outpainting operations, this particular use case has a strict set of requirements that we must uphold.
For example, when generating a background, any incidental modifications to the product itself could result in a potentially misleading advertisement.
However, since we can finetune a Canvas variant that directly targets this use case, we can design our outpainting pipeline to enforce these requirements and guarantee that our outputs are usable in production.
Our approach, diagrammed in Figure~\ref{fig:outpainting_inference_diagram}, incorporates these requirements both in the training of the model itself, as well as in additional filtering and post-processing applied at runtime.

\subsection{Dataset Collection and Training}
As previously stated, producing task-specific variants of Canvas largely centers around collecting training data that matches the desired use case, then fine-tuning the generic base model on this dedicated dataset.
In the case of background outpainting, we collect a dataset of in-context product photographs, where a product to be advertised is prominently featured in a real-world scene.

We use InSPyReNet~\cite{kim2022revisitingimagepyramidstructure} to obtain these foreground masks.
As a preprocessing step, we apply this mask to the input product photograph, removing the original background and replacing it with a plain white background.
Then, during training, the model is provided this masked product shot, along with a text caption describing the original image and background, and is supervised to reconstruct the original unmasked image.

During inference, after running the diffusion loop and obtaining the final generation, we then additionally composite the original product cutout back onto the generated image.
This is a further safeguard to avoid any potential modifications---our use of segmentation masks during training ensures the model won't extend the product beyond its existing boundaries, and reusing the original product shot ensures the model cannot inadvertently alter the product itself.
We also implement dynamic positioning and padding, deriving the vertical placement and padding from the original product shot instead of using a fixed centered layout, to better accommodate product-specific composition.

Aspect ratio outpainting follows a similar protocol, though naturally we no longer need to produce segmentation masks for the products themselves.
Thus, instead of masking out the background, we mask out bands along the top and bottom of the input images, and the model is trained to reconstruct the original image, thereby ``extending'' the training sample.

\subsubsection{Outpainting VAE}
\begin{figure}[t]                                       
  \centering                                                                   
  \includegraphics[width=0.90\linewidth]{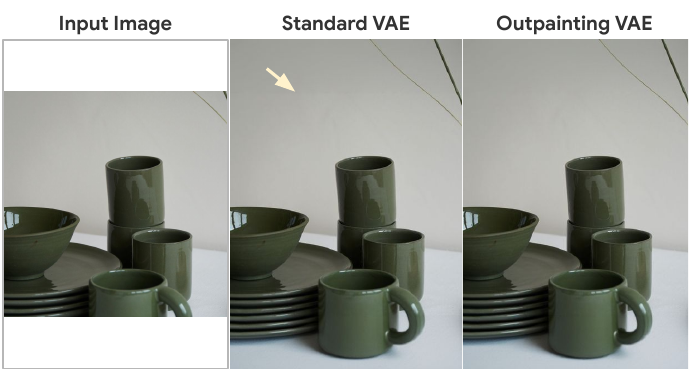} 
  \caption{Outpainting VAE comparison.}
  \Description{Outpainting VAE}
  \label{fig:outpainting_vae}                                               
\end{figure}

We observe occasional color mismatches, especially in the aspect ratio extension setting, where the boundary between the original image and the newly generated bands along the top and bottom are visibly distinct.
To address this issue, we start from the FLUX.1 VAE distilled on Pinterest data and fine-tune the decoder to accept mask and masked image inputs alongside the outpainted image. This improves color harmonization by blending the inputs across multiple feature scales. Figure~\ref{fig:outpainting_vae} compares decodes from the standard FLUX.1 VAE and our outpainting VAE using the same generated latents; the FLUX.1 VAE shows a visible background color discontinuity and seam at the boundary (marked by a yellow arrow), while the outpainting VAE produces a smoother blend across the transition.

\subsubsection{Super-resolution}
Our base outpainting pipeline generates images at $416 \times 608$ to improve the efficiency of multiple-generation settings.
To meet product high-resolution requirements, we upscale the top-ranked candidate by $\times3$ to $1248 \times 1824$ using a Canvas super-resolution variant, and then composite the original high-resolution product cutout back into the final output.
Notably, InSPyReNet segmentation is computed on the original high-resolution input to better capture the detailed product boundaries.

\subsection{Runtime Enhancements}
\subsubsection{Metaprompting in Background Outpainting}
We introduce a metaprompting layer that uses VLMs to produce diverse background prompts for each input product. These metaprompts target a range of background complexity from simple color backgrounds to lifestyle backgrounds so that we can increase the diversity across multiple generations and improve the likelihood of selecting a suitable prompt and producing a high-quality output.

\subsubsection{Eligibility}
Background outpainting is suitable for product images with plain white backgrounds, whereas aspect-ratio outpainting is primarily applied to lifestyle images with suboptimal aspect ratios. We therefore perform pre-generation eligibility filtering to focus on inputs that are likely to yield meaningful enhancement. In practice, we filter background outpainting candidates by requiring a white dominant background and no human presence. For aspect-ratio outpainting, we exclude white/solid-color backgrounds, require no human presence, and filter out images with a height:width aspect ratio $< 1.4$. We additionally exclude cases that are prone to failure modes e.g., out-of-frame products, heavy obstructions, overlays, text-dominant images, or patterned images.

\subsubsection{Human Review}
Merchants are often reluctant to deploy AI-generated content without reliable quality controls. Although VLMs offer a scalable and low-cost alternative for automated screening, we find that they do not consistently detect subtle defects or enforce strict requirements such as complete product preservation. We therefore design structured human-review templates and rely on trained human raters to identify fine-grained artifacts and ensure that outputs are safe and visually appealing. Each generated image is independently reviewed by two raters following the rubric and template in Appendix~\ref{app:appendix_human_rater_questions}. For background outpainting, raters assess product defects and background defects; for aspect-ratio outpainting, raters assess eligibility and background defects. An output is marked defective if either rater flags any defect. 

\subsubsection{Multiple-generation and Reward Model}
In our production pipeline, we use additional inference compute to produce multiple generations in order to increase the likelihood of obtaining an acceptable output.
When generating candidates, we try to diversify outputs where possible:
for background outpainting, diversity is introduced through different prompts, whereas for aspect-ratio outpainting it is introduced through different random seeds.
We then use a reward model to automatically select the single best candidate to send for rater review.
This reward model is trained by leveraging historical rater annotations, so that it aligns closely with rater preferences. Details of the reward model are in Appendix~\ref{app:details_of_reward_model}.

In practice, the dominant cost in our pipeline is not compute or latency but the cost of human review, meaning that generating more candidates and selecting the best is a cost-effective way to increase the volume of successful generations without increasing the human review workload.
We found that using even just two candidates was enough to realize tangible improvements in our yield rate.
To further minimize the runtime impact, candidate generation occurs at the lower resolution ($416{\times}608$), and only the selected candidate undergoes superresolution; as a result, generating two candidates instead of one increases overall runtime by only $\sim$20\% compared to the single-candidate baseline.
We also evaluated generating more candidates per image but observed diminishing returns---for example, generating eight candidates instead of two only increases relative yield from 7\% to 10\%.

\subsubsection{Seed Tuning}
We find that the seed used for diffusion noise initialization can significantly influence output quality for both backdrop and aspect-ratio outpainting, as consistent with prior work~\cite{xu2025goodseedmakesgood}.
To select good seeds for our specific tasks, we run an offline sweep by generating outputs for 100 inputs across 1,024 seeds, then ranking seeds by the average reward attained by their outputs.
When running the model in production, we fix the seed used during generation to the highest ranking subset.

\subsubsection{Post-generation Filtering}
In some cases, even the top-ranked candidate receives a low reward score due to predictable failure modes, e.g., segmentation errors, out of frame products, etc. This indicates that the product input is unlikely to pass human review and should be filtered out before review. To address this, we generate a larger batch of candidates, score and rank them with the reward model, and discard the product inputs with severe predicted defects that are likely to fail in subsequent human evaluations, thereby improving the overall yield.

\section{Experiments}
In this section, we first validate our shared base model design, then compare Canvas against several third-party models for background outpainting using offline human evaluations. We further show that Canvas delivers metric wins in online A/B tests for both background outpainting and aspect-ratio outpainting. We also showcase other fine-tuned Canvas variants, including multi-image scene synthesis and image-to-motion generation.

\subsection{Effect of Task-Specific Fine-Tuning}
We begin by examining the effect of task-specific fine-tuning on performance, using background outpainting as our target task. The shared base model, evaluated before dedicated fine-tuning, achieves an overall no-defect rate of 54.2\%, compared to 58.4\% for the task-specific fine-tuned variant (N=637), confirming that fine-tuning leads to a material gain. 
We also compared against a previous Canvas version trained solely on the two outpainting tasks, without any additional tasks; that model achieves a no-defect rate of 58.9\%, within error of our fine-tuned result, suggesting that sharing a base model across additional tasks does not meaningfully hurt downstream performance. Additionally, sharing a base model reduces training cost by 17\% compared to training separate models for the two outpainting tasks, and becomes increasingly cost-effective as more use cases are added.

\subsection{Offline Evaluations for Outpainting}
\begin{figure}[t]
  \centering
  \includegraphics[width=0.93\linewidth]{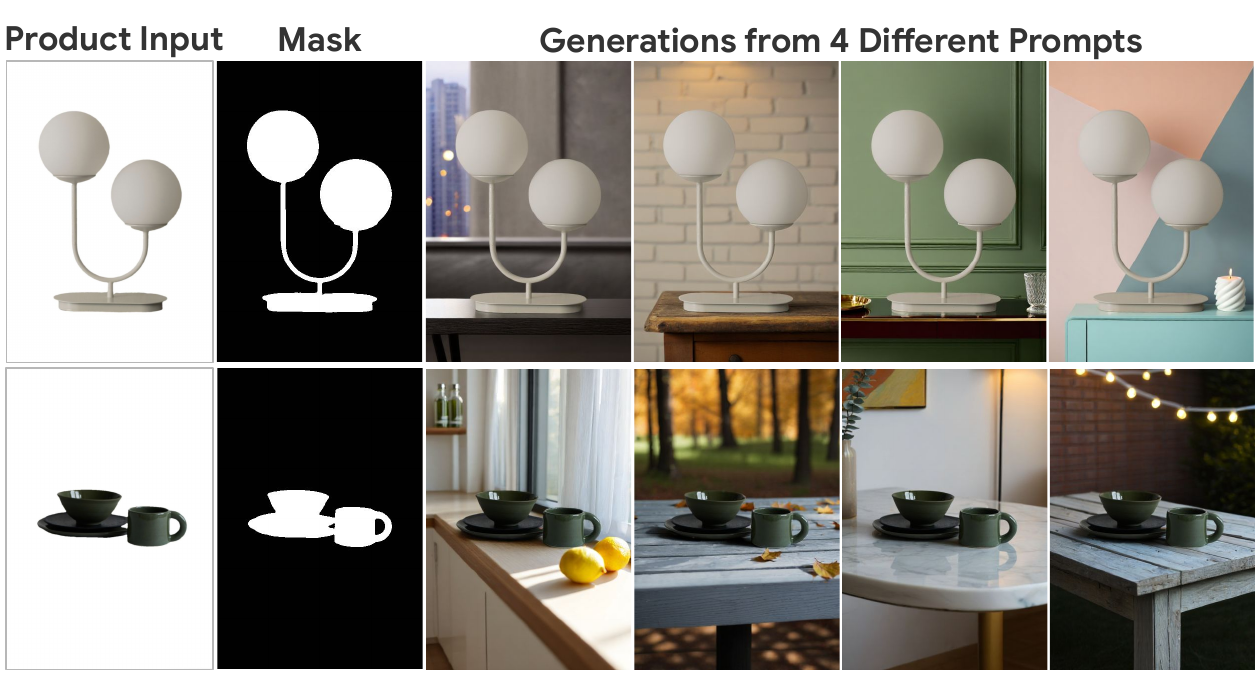}
  \caption{Visual examples for background outpainting.}
  \Description{Visual examples for background outpainting}
  \label{fig:result_background_outpainting}
\end{figure}

\begin{figure}[t]
  \centering
  \includegraphics[width=0.93\linewidth]{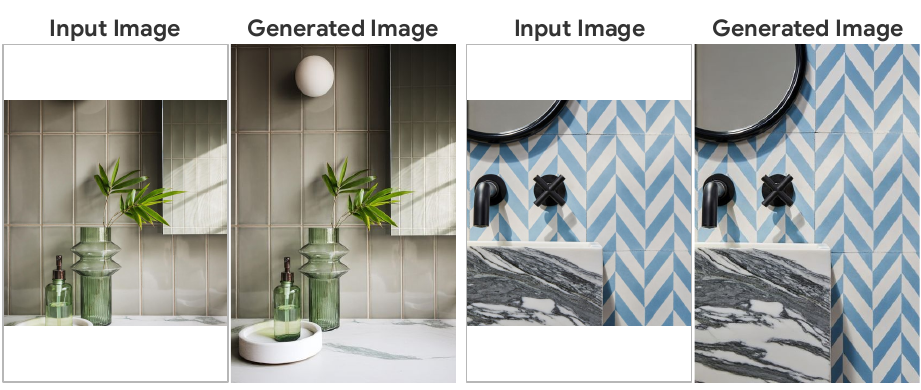}
  \caption{Visual examples for aspect-ratio outpainting.}
  \Description{Visual examples for aspect ratio outpainting}
  \label{fig:result_aspect_ratio_outpainting}
\end{figure}

We show two examples in Figure ~\ref{fig:result_background_outpainting}, each with four outpainted backgrounds generated from different prompts. Our model produces high-fidelity results that strictly preserve the product identity while integrating it naturally into a high-quality, more engaging background than the original white backdrop.
Examples before and after expanding to 3:2 aspect-ratio are shown in Figure ~\ref{fig:result_aspect_ratio_outpainting}.

We compared Canvas with GPT-Image, FLUX.1 Kontext and Google's Nano Banana on the background outpainting task over $N{=}996$ products in Table ~\ref{tab:human-eval-bg-outpainting}. All methods are evaluated using the same prompt for the same product input, and for fair comparison, candidate generation and reward model ranking was not used for Canvas. Each generated image is independently evaluated by two trained human raters using the template in Appendix~\ref{app:appendix_human_rater_questions}. We also report an overall no-defect rate, where an output is considered defective if either rater identifies a product or a background defect.

\begin{figure}[t]
  \centering
  \includegraphics[width=0.95\linewidth]{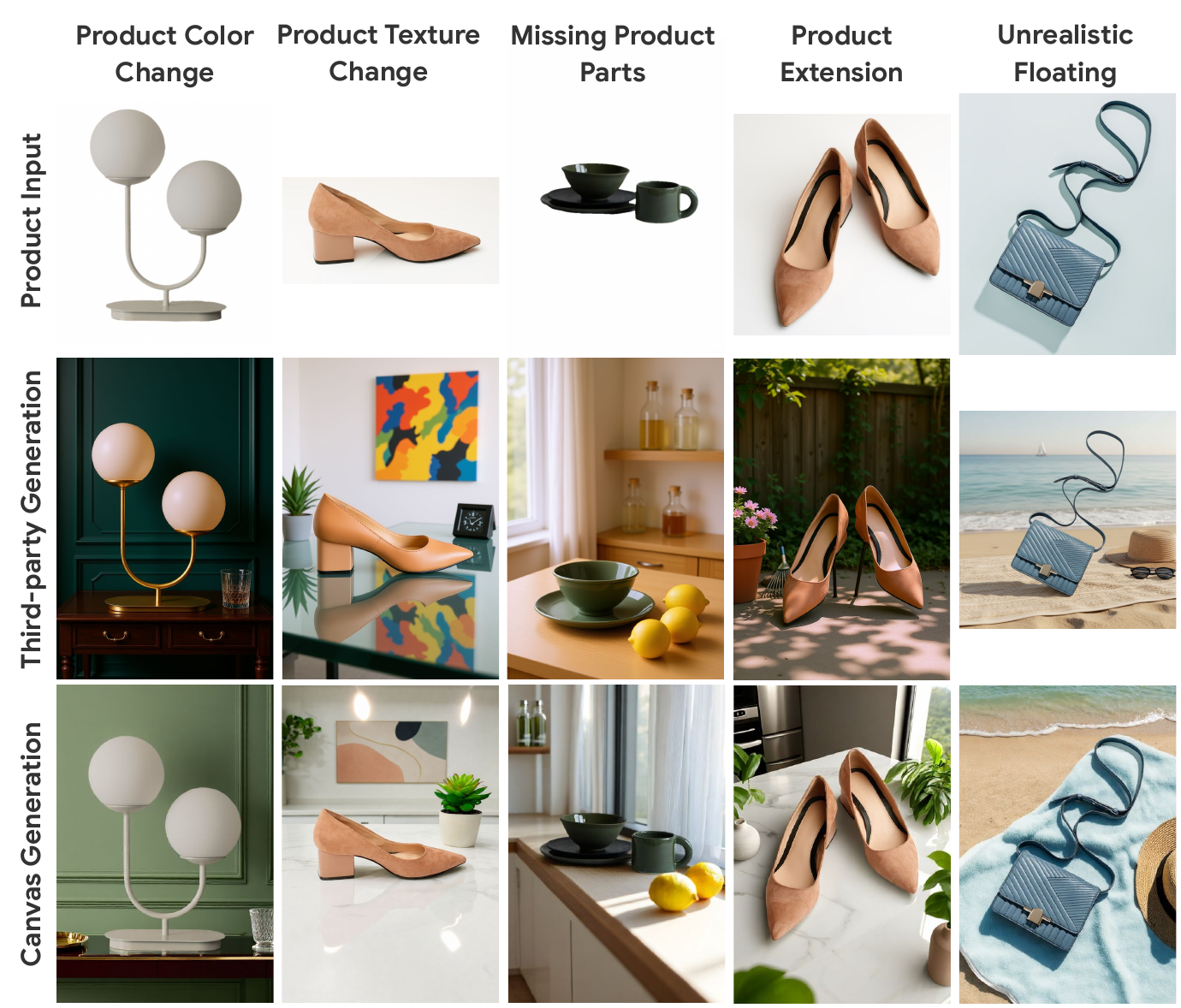}
  \caption{Failure cases of third-party models.}
  \Description{Failure Cases of Third-party Models}
  \label{fig:failure_cases_third_party_models}
\end{figure}

A common failure of third-party models is imperfect product preservation such as changes in color or product extensions, which results in relatively high product defect rates. Canvas preserves product identity significantly better with fewer defects, leading to a higher overall yield rate. Some failure cases of third-party models are shown in Figure~\ref{fig:failure_cases_third_party_models}.

\begin{table}[t]
  \caption{Human evaluation for background outpainting.}
  \label{tab:human-eval-bg-outpainting}
  \centering
  \begin{tabular}{lccc}
    \toprule
    & \multicolumn{3}{c}{No-defect Rate}\\ \cmidrule(l){2-4}
    Model &
    Product & Background & Overall \\
    \midrule
    GPT-Image 1.5 & 52.9\% & 50.8\% & 26.2\% \\
    FLUX.1 Kontext & 53.4\% & 55.8\% & 28.2\% \\
    Nano Banana & 74.6\% & 55.9\% & 42.5\% \\
    Canvas & 84.0\% & 54.9\% & 47.2\% \\
    \bottomrule
  \end{tabular}
\end{table}

\subsection{Online A/B Results for Outpainting}
While integrating Canvas into the Pinterest Performance+ suite, we conducted A/B tests to measure its effect on user engagement. For background outpainting, the control group contained the original product shot images with white backgrounds, and the treatment group contained images with backgrounds generated by Canvas on the Home Feed, Related Pins, and Search surfaces. For aspect-ratio outpainting, the control group contained original product shot images with lifestyle backgrounds, and the treatment group contained images extended vertically by Canvas.
Detailed click and impression counts for both A/B test arms, along with production traffic volumes, are provided in Appendix~\ref{app:ab_test_details}.

The results of this experiment are shown in Table ~\ref{tab:engagement_outpainting}.
Images enhanced by Canvas consistently yielded higher engagement metrics across click-through rate (CTR), good clicks lasting 30 seconds (gCTR30), and click volume, often by double digit percentages.  
\begin{table}[t]
  \caption{
    Engagement results for outpainting use-cases.
  }
  \label{tab:engagement_outpainting}
  \centering
  \begin{tabular}{lcc}
    \toprule
    Metric & 
    Background &
    Aspect-ratio \\
    \midrule
    CTR  & +18.0\% & +12.5\% \\
    gCTR30 & +7.6\% & +6.8\% \\
    Click Volume & +18.6\% & +12.9\% \\
    \bottomrule
  \end{tabular}
\end{table}

\subsection{Additional Canvas Variants}

\begin{figure}[t]
  \centering
  \includegraphics[width=0.95\linewidth]{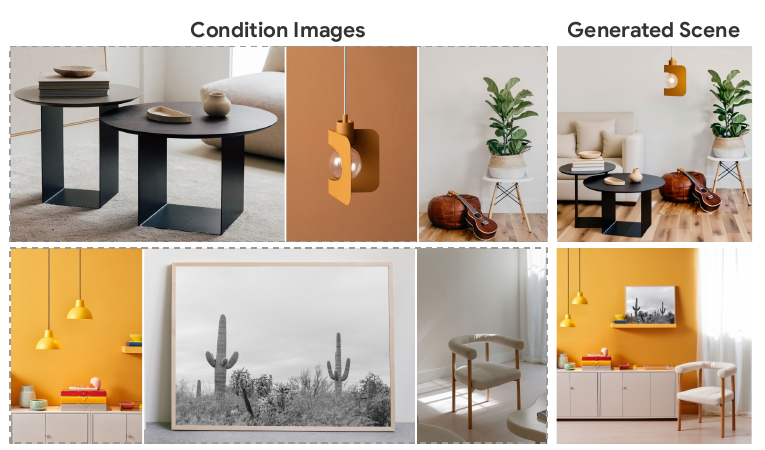}
  \caption{Visual examples for multi-image scene synthesis.}
  \Description{Visual examples for multi-image scene synthesis}
  \label{fig:result_moss}
\end{figure}

\paragraph{Multi-Image Scene Synthesis}
To allow for more flexibility in visualization than that afforded by outpainting, including viewpoint and lighting transformations, as well as allowing for multiple image inputs, we conducted initial explorations of multi-image scene synthesis in the home decor vertical.
In a straightforward extension of Canvas training, we move from sequence-concatenating one VAE latent to up to eight VAE latents, depending on the number of products.
Using scenes as ground truth supervision, we fine-tune Canvas on our multi-image scene synthesis dataset. We found Canvas generalized well to multi-image inputs, incorporating input subjects and styles into a cohesive and aesthetically pleasing scene, while preserving object identity. In Figure~\ref{fig:result_moss}, we provide qualitative examples with input products spanning multiple home decor categories.

\paragraph{Image-to-Motion Generation}

\begin{figure}[t]
  \centering
  \includegraphics[width=0.95\linewidth]{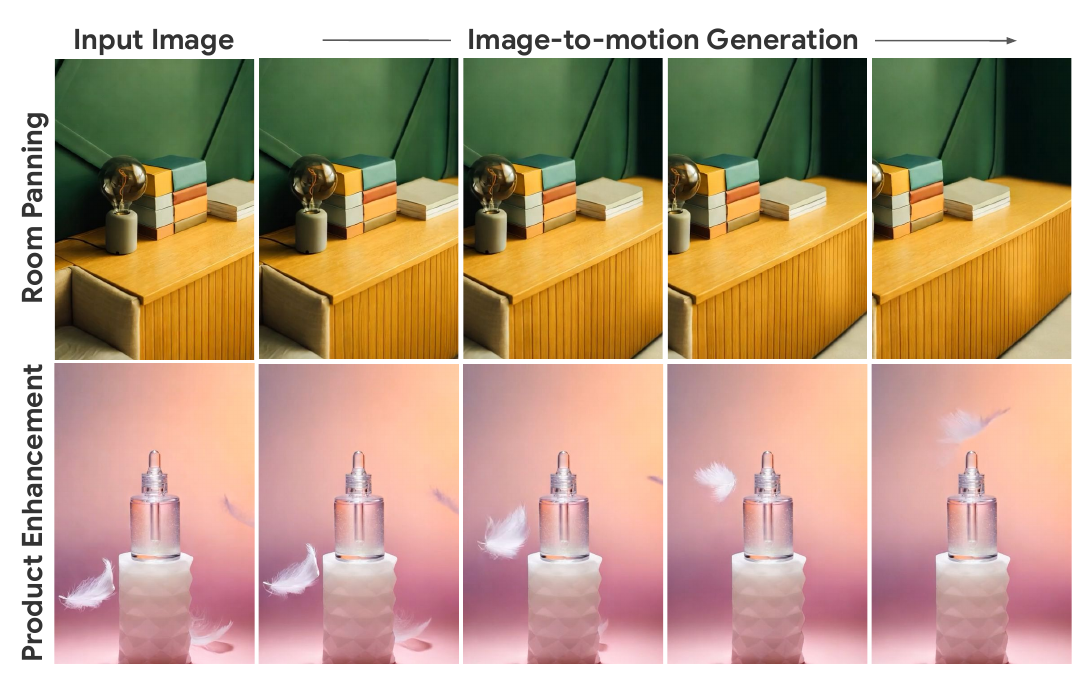}
  \caption{Visual examples for image-to-motion generation.}
  \Description{Visual examples for image-to-motion generation}
  \label{fig:result_video}
\end{figure}

In addition to single-image generation, we ran a pilot study on image-to-motion generation by fine-tuning Canvas with short video clips. Starting from our pretrained DiT backbone, we adapt the model to a video-compatible setting by using the Wan 2.2 VAE ~\cite{wan2025} as the latent representation, combined with 3D RoPE ~\cite{su2021_roformer} to encode spatio-temporal positions. Unlike narrative-style video generation, our target application is brief (2s) motion synthesis that adds dynamic enhancement to the input image. We found that our base model pretrained on multimodal image editing data provides an effective initialization for this task. In Figure~\ref{fig:result_video}, we provide qualitative examples for room-panning and product-background-enhancement motion generation.

\vspace{6pt}
In addition to the use cases shown above, more product-related examples generated from base Canvas are provided in Appendix~\ref{app:more_product_editing_examples}.

\section{Conclusion}
This paper presents Pinterest Canvas, a large-scale image generation system developed at Pinterest to support image editing and enhancement use cases.
Our core design is as follows: train a base model with broad editing capability, then rapidly fine-tune task-specific variants on dedicated datasets with targeted strategies to better satisfy individual product requirements.
We also share practical learnings from building Canvas, including dataset curation, training, and inference for the base model, as well as the outpainting pipeline and additional strategies used to meet specific product requirements. Offline comparisons against third-party models and online experiments on outpainting use-cases show that Canvas showed superior performance with fewer defects and improvements in engagement metrics. 
Finally, we showcase additional variants such as multi-image scene synthesis and image-to-motion generation, demonstrating the flexibility and generalization capabilities of base Canvas to rapidly fine-tune to specific product use-cases.

For zero-shot, open-world image generation, large generic models often remain the best option.
However, when building specific products with strict requirements, we recommend starting with a strong foundational model and fine-tuning on data that directly matches the target use case, rather than relying on brittle approaches like prompting or input formatting or training multiple task-specific models from scratch. This gives more direct control over model behavior and performance and enables rapidly onboarding new use cases, while sharing most training compute burden across all task models. We encourage others to explore this approach for deploying generative models to production with confidence.

\begin{acks}
We want to thank Tiffany Chen for her help with the third-party model comparisons and her work on Canvas; Blake Lewis, Shubham Gupta, Misha Tsysin, Przemyslaw Ziaja, Eugene Zabrotsky, Natasha Karmali, Teddy White and Ryan Galgon for their contributions to Canvas; Tianyuan Cui, Nestor Basterra, Enbo Zhou and Jason Shu from Pinterest Ads team for their partnership; David Xue, Ric Poirson and Rohan Mahadev for their multimodal dataset collection effort.
\end{acks}

\bibliographystyle{ACM-Reference-Format}
\bibliography{sample-base}

\appendix

\section*{Appendix}

\section{Details of Training Dataset}
\label{app:details_training_dataset}
\paragraph{Multi-view product.}
We curate a multi-view product dataset consisting of sets of images from the same item, sourced from the Pinterest Unified Shopping Catalog. We apply a VLM-based verification step to assess whether images within each set correspond to the same product and to remove mismatched items or product variants. This dataset improves model abilities of multi-view object revisualization and object identity preservation.

\paragraph{OmniSage neighbors data.}
OmniSage~\cite{Badrinath_2025} is a heterogeneous graph that models neighbor relationships among Pinterest entities such as Boards and Pins. We construct training pairs by retrieving OmniSage neighbors across all verticals. For each anchor Pin image, we sample a related neighbor, use VLMs to describe the relationship between the two images and generate a detailed caption to reproduce the target image from the input, and format each example as $(\text{Pin} + \text{modifier/instruction}) \rightarrow \text{neighbor}$. We additionally use Pinterest's internal image embedding model, PinCLIP~\cite{Beal2026PinCLIP}, to compute cosine similarity between the two images and filter candidate pairs with scores that are too low or too high, removing both near-duplicates and weakly related matches.

\paragraph{Background outpainting.}
We start from the Pinterest Unified Shopping Catalog and promoted Pins, filter to the Stock Photo catalog (lifestyle images with a primary product), and then deduplicate the data. We further remove copyright-restricted content as well as crawled and collage images. We use VLMs to generate synthetic captions describing both the product and its surrounding background, and then produce salient-object segmentation masks using InSPyReNet~\cite{kim2022revisitingimagepyramidstructure}.

\paragraph{Aspect-ratio outpainting.}
The goal of this task is to extend an image vertically to a 3:2 (height:width) aspect ratio. We synthesize the aspect-ratio dataset from our text--image data by applying random height masking to the original image and masking out the top and bottom regions to be generated.

\paragraph{Super-resolution.}
The goal of this task is to perform $\times3$ super-resolution restoration. We synthesize training pairs from our text--image data using ESRGAN~\cite{wang2018esrganenhancedsuperresolutiongenerative} degradations.

\paragraph{Product extraction and Scene synthesis.}
We collect training samples consisting of pairs of images of a single product.
These images are chosen such that one image is a product shot, and the other image is of a scene containing that product.
These matches are initially collected using our Shopping Catalog, then verified and captioned with a VLM to ensure that the images contain the same product.
Depending on which image we use as the input condition and which one we use as the target image, we effectively form samples that train the model to perform two related tasks: either product extraction (generate a product shot from an in-context view) or scene synthesis (from a product shot, generate a scene containing that product).

\paragraph{Multi-image scene synthesis.}
We curate a multi-image scene synthesis dataset consisting of a scene paired with a set of one or more product images. Starting from scenes that have been tagged with products by Pinterest users, we verify that the referenced product images are exact matches using a VLM. We apply internal image quality filters to the scenes to steer the model toward high quality outputs, and remove near-duplicate scene-product pairs. We use a VLM to caption the scenes, describing both the overall setting and the placement of prominent objects, and use these captions as text conditions.

\paragraph{Video-augmented key frame pairs.}
We additionally introduce frame pairs from videos in the Pinterest corpus, to learn temporally consistent transformations. Each video is first segmented into short, coherent clips, which are then filtered using both an optical-flow–based method and VLMs to remove near-static clips and flickering clips. From each remaining clip, we select the first frame and a second frame with a minimum temporal separation of 2 seconds. The transitions between the two frames may capture global changes (e.g. camera panning) or local changes (e.g. object moves while camera is static). We use VLMs to describe the transformation and use that as the text supervision.

\section{Outpainting Human Review Questions}
\label{app:appendix_human_rater_questions}
We developed separate human-review questions and rating guidelines for background outpainting and aspect-ratio outpainting to identify defects in generated images and ensure compliance with product requirements. 

\begin{figure}[!t]
  \centering
  \includegraphics[width=\linewidth]{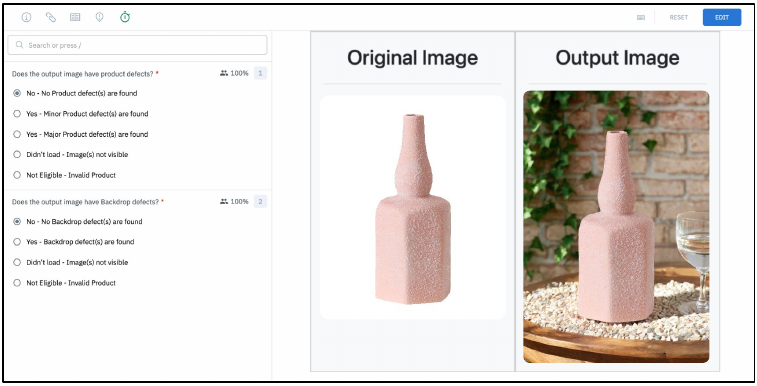}
  \caption{Human review questions for background outpainting.}
  \Description{Human review questions for background outpainting.}
  \label{fig:human_review_questions_background_outpainting}
\end{figure}

For background outpainting, raters first view the original image for around 5 seconds to identify the main product and any supporting elements in the image. They then view the generated output image for another 5 seconds, assess whether any defects occurred to the main product, then note any other defects occurring in the rest of the image. Finally, they answer the two questions shown in Figure~\ref{fig:human_review_questions_background_outpainting}.

\begin{figure}[!t]
  \centering
  \includegraphics[width=\linewidth]{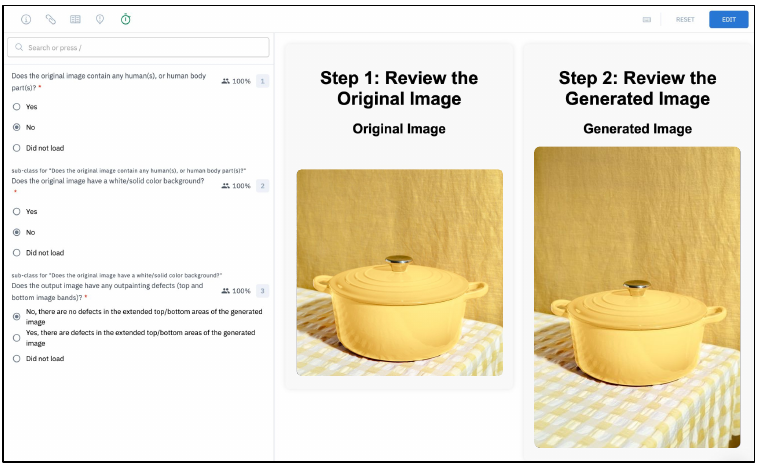}
  \caption{Human review questions for aspect-ratio outpainting}
  \Description{Human review questions for aspect-ratio outpainting}
  \label{fig:human_review_questions_aspect_ratio_outpainting}
\end{figure}

For aspect-ratio outpainting, raters first view the left original image and visually capture the details of the overall image and its top and bottom edges. They then confirm whether any humans or human body parts are present and answer Question 1 as in Figure~\ref{fig:human_review_questions_aspect_ratio_outpainting}. Next, they determine whether the original image has a white or solid-color background and answer Question 2. Raters then view the right outpainted image, focusing on the top and bottom image bands and answer Question 3. The first two questions serve as a human eligibility filter to exclude images with human presence and to remove cases that are unlikely to benefit from aspect-ratio outpainting, and Question 3 focuses on defect detection, helping filter out any outputs that contain visible artifacts or other quality issues.

\FloatBarrier
\section{Details of Outpainting Reward Models}
\label{app:details_of_reward_model}
To train outpainting-specific reward models that can accurately predict defects, we train two separate reward models for background outpainting and aspect-ratio outpainting using the corresponding large-scale retrospective human review annotations. We use a Vision Transformer (ViT) as the reward model backbone. For training loss, we adopt an aggregated rater probability loss, which captures the likelihood that an image contains defects, rather than treating each rater judgment as an independent sample. This avoids conflicting supervision (e.g., the same image labeled both defective and non-defective by different raters) and makes the absolute value of reward scores more interpretable to better reflect the severity of defects, so we can use the absolute reward value for post-generation eligibility filtering. For the model inputs, we increase the input resolution from a $224 \times 224$ square-padded format to $416 \times 608$ to better match the generated outputs and capture fine-grained details. In addition, we feed the original and generated images as a paired, two-channel input, and find that using the original image as a reference baseline improves defect-detection performance.

\section{Outpainting A/B Test Details}
\label{app:ab_test_details}
For background outpainting, the control arm received 117.0K clicks over 16.4M impressions, while the treatment arm received 138.7K clicks over 16.5M impressions. For aspect-ratio outpainting, the control arm received 123.6K clicks over 10.7M impressions, while the treatment arm received 139.6K clicks over 10.7M impressions. In production, daily impressions for Canvas-generated images are approximately 75M for background outpainting and 15M for aspect-ratio outpainting.

\FloatBarrier
\section{Visualizations of More Product Editing Examples}
\label{app:more_product_editing_examples}
\begin{figure}[!t]
  \centering
  \includegraphics[width=1.0\linewidth]{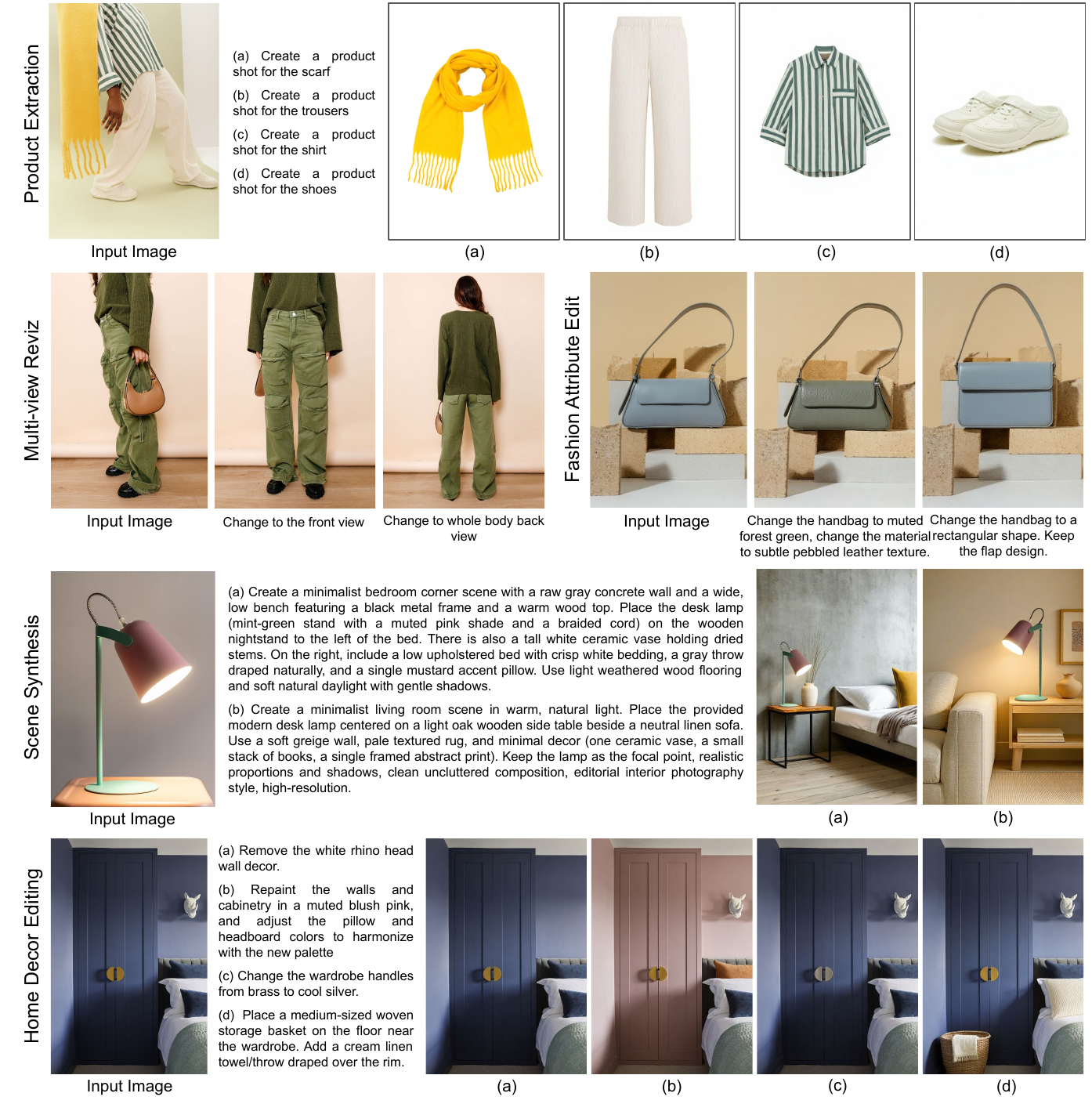}
  \caption{Visualizations of more product editing examples}
  \Description{More product use-case examples}
  \label{fig:more_product_editing_examples}
\end{figure}

We show more product-related image-editing applications using base Canvas in Figure~\ref{fig:more_product_editing_examples}, e.g. product extraction, multi-view revisualization, fashion attribute edit for fashion images, and single-image scene-synthesis and home decor editing for home decor images.

\end{document}